\documentclass[10pt,twocolumn,letterpaper]{article}

\usepackage{cvpr}              % To produce the CAMERA-READY version
\usepackage[accsupp]{axessibility}
\usepackage{algorithm}
\usepackage{algorithmic}
\usepackage{tabularray}
\usepackage{amssymb}
\usepackage{multirow}
\usepackage{booktabs}
\usepackage{float}
\usepackage[dvipsnames]{xcolor}

\definecolor{cvprblue}{rgb}{0.21,0.49,0.74}
\usepackage[pagebackref,breaklinks,colorlinks,citecolor=cvprblue]{hyperref}

\def\paperID{8676} % *** Enter the Paper ID here
\def\confName{CVPR}
\def\confYear{2025}

\title{Federated Learning with Domain Shift Eraser}

\author{
    Zheng Wang\textsuperscript{1,2,3} \quad Zihui Wang\textsuperscript{5} \quad Zheng Wang \textsuperscript{4} \quad  Xiaoliang Fan \textsuperscript{1,2 } \quad Cheng Wang \textsuperscript{1,2}\thanks{Corresponding Author}\\
    \textsuperscript{1}Fujian Key Laboratory of Sensing and Computing for Smart Cities, Xiamen University, China \\
    \textsuperscript{2}Key Laboratory of Multimedia Trusted Perception and Efficient Computing,\\
Ministry of Education of China, School of Informatics, Xiamen University, China \\
    \textsuperscript{3} Shanghai Innovation Institution\\
    \textsuperscript{4} School of Informatic, Xiamen University, China\\
    \textsuperscript{5} Peng Cheng Laboratory, Shenzhen, China
}
\begin{document}
\maketitle
\begin{abstract}
Federated learning (FL) is emerging as a promising technique for collaborative learning without local data leaving their devices. However, clients' data originating from diverse domains may degrade model performance due to domain shifts, preventing the model from learning consistent representation space. In this paper, we propose a novel FL framework, Federated Domain Shift Eraser (FDSE), to improve model performance by differently erasing each client's domain skew and enhancing their consensus. First, we formulate the model forward passing as an iterative deskewing process that extracts and then deskews features alternatively. This is efficiently achieved by decomposing each original layer in the neural network into a Domain-agnostic Feature Extractor (DFE) and a Domain-specific Skew Eraser (DSE). Then, a regularization term is applied to promise the effectiveness of feature deskewing by pulling local statistics of DSE's outputs close to the globally consistent ones. Finally, DFE modules are fairly aggregated and broadcast to all the clients to maximize their consensus, and DSE modules are personalized for each client via similarity-aware aggregation to erase their domain skew differently. Comprehensive experiments were conducted on three datasets to confirm the advantages of our method in terms of accuracy, efficiency, and generalizability.

% In this paper, we introduce Federated Prototypes Learning (FPL) to tackle the challenges posed by domain shifts in federated learning. The central concept involves the creation of cluster prototypes and unbiased prototypes, which offer valuable domain insights and establish a fair target for convergence. Specifically, we aim to bring the sample embeddings closer to the cluster prototypes that share the same semantic meaning, while simultaneously distancing them from prototypes of different classes. Additionally, we implement consistency regularization to align local instances with their corresponding unbiased prototypes. Our experimental results on the Digits and Office-Caltech datasets highlight the effectiveness of our proposed approach and the efficiency of its key components. 

% The ABSTRACT is to be in fully justified italicized text, at the top of the left-hand column, below the author and affiliation information.
% Use the word ``Abstract'' as the title, in 12-point Times, boldface type, centered relative to the column, initially capitalized.
% The abstract is to be in 10-point, single-spaced type.
% Leave two blank lines after the Abstract, then begin the main text.
% Look at previous \confName abstracts to get a feel for style and length.
\end{abstract}

\section{Introduction}
\label{sec:intro}

\begin{figure}
    \centering
    \includegraphics[width=\linewidth]{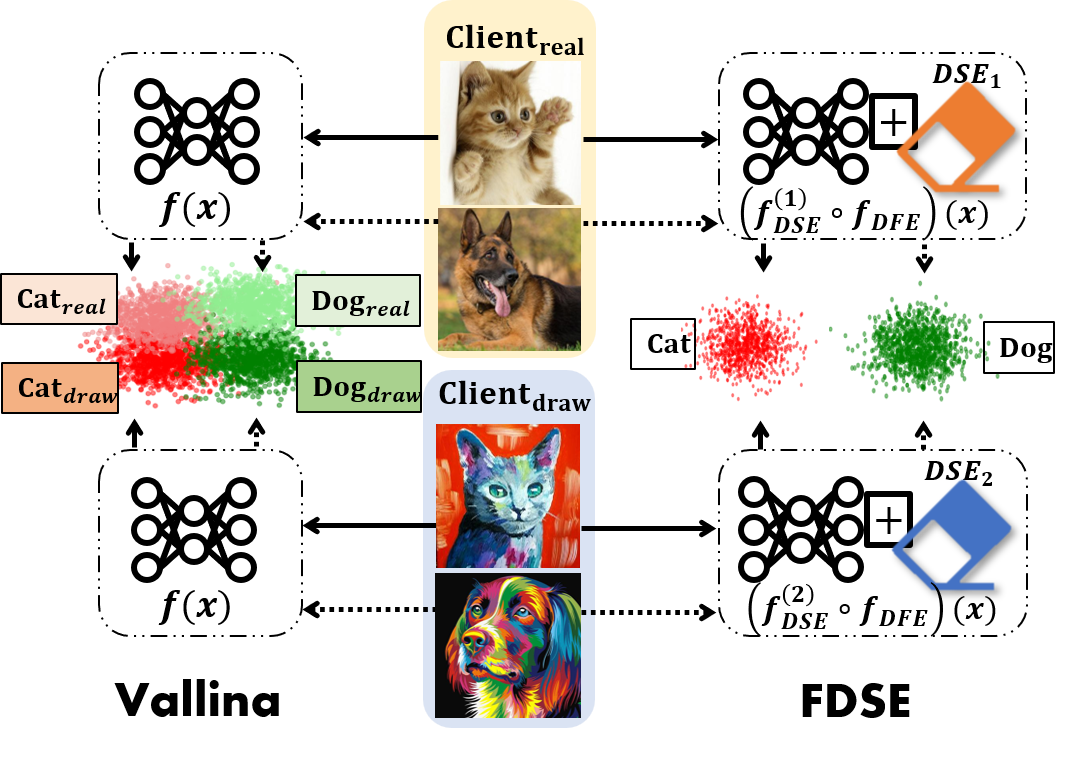}
    \caption{\textbf{Inllustration of the characteristic of the proposed FDSE v.s. Vallina FL \cite{mcmahan2017communication}}. FDSE decomposes the model to respectively erase domain skew for each client by $f_{DSE}^{(i)}$ and extract generalizable features for all the clients by $f_{DFE}$. This promotes consistency in representation space across domains since knowledge is fine-grainedly decoupled for learning and aggregation.}  
    \label{fig1}
    \vspace{-0.5cm}
\end{figure}

Federated learning has emerged as a promising paradigm for training machine learning models on distributed data while preserving privacy \cite{mcmahan2017communication}. However, a notable challenge arises when clients originate from diverse domains, where domain shifts across clients may result in performance degradation of the aggregated model \cite{huang2023rethinking,chen2020fedhealth}. For example, hospitals located in different regions may collect data from diverse patient populations using various machines and protocols, leading to distinct domains in their feature distributions \cite{li2021fedbn}. 
The misalignment in feature spaces can hinder client consensus on the representation space (e.g., Figure \ref{fig1} left), finally reducing model performance in two ways. First, the model is compelled to additionally learn generalizable representations for vastly different samples across domains, competitively preventing it from focusing on the task objective. Second, the training process may be dominated by a single domain when clients’ model updates exhibit conflicts and substantial variations in magnitude \cite{chen2024fair,wang2021fedfv}, which causes the model to overfit on the prevailing domain while sacrificing its utility on others \cite{hu2022federated}.

% On one hand, the model is forced to learn generalizable representations of different domains' samples that can often be vastly disparate, thereby competitively preventing the model from focusing on the task objective itself. On the other hand,  the training process may be dominated by a single domain when clients' model updates exhibit conflicts and significant variations in magnitude \cite{chen2024fair,wang2021fedfv}, causing model overfitting on the dominant domain while sacrificing its utility on other domains \cite{hu2022federated}. 
% The misalignment in feature spaces can degrade model performance in two ways. First, it forces the model to learn generalizable representations that can be adapted to various domains. This poses a great challenge for the model to align different domains' features that can often be vastly disparate, thereby competitively preventing the model from focusing on the task objective itself. Second, the training process may be dominated by a single domain when clients' model updates exhibit conflicts and significant variations in magnitude \cite{chen2024fair,wang2021fedfv}. The lack of consensus can result in model overfitting on the dominant domain while sacrificing its utility on other domains \cite{hu2022federated}. 

Recent works addressing the FL domain shift problem can be categorized into two main groups. The first group (i.e., consensus-based methods) \cite{zhou2023fedfa,tan2022fedproto, chen2024fair} enhances the consensus among clients at various levels to improve the model generalizability across domains. In contrast, another group (i.e., personalization-based methods) \cite{li2021fedbn,liang2020think, sun2021partialfed} mitigates the need for clients' consensus by personalizing their models to fit their local data distributions.
Since the two groups exhibit complementary strengths in modeling different types of knowledge (i.e., general and personal knowledge), it's natural to consider a hybrid solution that synergizes their advantages for further improvement. However, model personalization may severely hinder clients' consensus due to over-focusing on clients' local benefits, thereby resulting in invalid integration. This motivates us to rethink the FL domain shift from the hybrid perspective
\begin{center}
      \textit{How to personalize models to enhance clients' consensus?}
\end{center}
% Despite the rapid development of each group of methods, a hybrid solution that synergizes their advantages remains unexplored, leading to a potentially large space for further model performance enhancement. 
% a hybrid solution that integrates their advantages is commonly overlooked, leaving a potentially large space for further model performance enhancement. To bridge this gap, we rethink the FL domain shift problem from the hybrid perspective: 

%  The inherent conflict in 
% Despite the rapid development of each group of methods, a hybrid solution that synergizes their advantages remains unexplored, leading to a potentially large space for further model performance enhancement. One possible reason lies in that To bridge this gap, we rethink the FL domain shift problem from the hybrid perspective: 
% For example, [FedFA][FedProto][FPL] maintains consistent embeddings extracted by models trained on different domains' data, while FedHEAL aggregates model updates to be consistently beneficial to all the clients. 

% For example, FedBN and FedDAR respectively keep batch normalization layers and the output layers for each domain to explicitly handle different domain information.

In this work, we address this issue by \textbf{using personalized parameters to erase domain-specific skew for each client while enhancing clients' consensus on other parameters} based on two key observations. First, visual learning tasks often mitigate samples' domain-specific skew differently. For example, in point cloud segmentation, high-density and low-density point clouds from channel-varying lidars are respectively downsampled and upsampled to achieve their alignment in feature space \cite{yuan2024density}. These inherently opposing processes of feature alignment necessitate personalized modeling.
% Utilizing a set of model parameters to learn opposing operations will lead to increased competition with the task objective itself.
Second, a unified treatment is usually applied to different samples after their domain-specific skew is mitigated \cite{jiang2022harmofl, yuan2024density, huang2023rethinking}, revealing the importance of achieving post-deskewing consensus in a domain-agnostic way.
% For example, a unified classifier is commonly used to predict the classes of samples from different domains after they are projected into a consistent representation space. 
% Besides, this paradigm naturally decouples the two objective 
% following the two principles: 1) different domain skew should be eliminated in different ways (i.e. \textbf{domain-specific skew elimination}), and 2) the deskewed features from different domains should be fairly treated by the model (i.e., \textbf{domain-agnostic fair treatment})
Following the above paradigm, we develop a novel FL framework, \textbf{F}ederated \textbf{D}omain \textbf{S}hift \textbf{E}raser (\textbf{FDSE}), aiming to differently erase clients' domain-specific skew and then enhance their consensus in a decoupled manner. 
 First, we formulate the model forward passing as an iterative deskewing process that extracts and then deskews features alternatively. This is efficiently achieved by decomposing each original layer in the neural network into a Domain-agnostic Feature Extractor (DFE) and a Domain-specific Skew Eraser (DSE). Then, a regularization term is applied to enhance the effectiveness of feature deskewing by pulling local statistics of each DSE's output close to the globally consistent ones.
 Finally, DFE modules are fairly aggregated and broadcast to all the clients to maximize their consensus, while DSE modules are respectively aggregated for each client based on their similarity to erase heterogeneous domain skew collaboratively. 
 % Further, we enhance the effectiveness of deskewing features by regularizing the model to pull local statistics of all DSE's outputs close to the globally consistent ones. 
% First, we view 
% alternatively extract domain-agnostic features and then erase domain-specific skew layer-by-layer during the model forward passing. Specifically, we separate each layer in the model into a Domain-agnostic Feature Extractor (DFE) and a Domain-specific Skew Eraser (DSE) (e.g., Figure \ref{fig1}), which extracts and then deskews features layer-wisely. Then, each GFE module is fairly aggregated among all the clients to enhance their consensus, while each DSE module is only shared among similar clients to erase domain-specific skew in a personalized way. Further, we regularize the local statistics of each DSE's outputs to be close to its corresponding global statistics, promoting consistency in the deskewed features from different domains.
Compared to the Vallina FL in Figure \ref{fig1} (left), our FDSE benefits from the design that emphasizes fine-grained personalization and consensus maximization on decoupled objectives, promoting both the consistency in the representation space and the efficiency of aggregating different types of knowledge. Our contributions are summarized as follows:
\begin{itemize}
    \item We rethink the domain shift problem in FL from a novel hybrid perspective that integrates the advantages of two distinct sets of existing methods, i.e., consensus-based and personalization-based methods.
    \item We develop a novel framework, FDSE, to improve cross-domain FL by differently erasing clients' domain skew while enhancing their consensus in a decoupled manner.
    % enhance client consensus upon de-skewed cross-domain features, which combines the advantages of two mainstream groups of existing methods  (i.e., model personalization and consensus maximization).
    \item We conduct comprehensive experiments on three datasets (i.e., Office-Caltech10, PACS, and DomainNet) to confirm FDSE's superiority in terms of accuracy, efficiency, and generalizability against the state-of-the-art methods.
\end{itemize}

\section{Related Works}
\label{sec:related}
\subsection{Data Heterogeneity in FL}

FL often faces challenges due to data heterogeneity, complicating the training process and degrading overall model accuracy \cite{mcmahan2017communication, zhao2018federated}. Previous efforts have been devoted to handling heterogeneous data by maintaining consistency between local objective and global objective \cite{li2020federated, li2021model, acar2021federated, karimireddy2020scaffold} or smoothening the clients' loss landscape during model training \cite{sun2023fedspeedlargerlocalinterval, mendieta2022local, qu2022generalized, dai2023fedgamma}.  Another series of works instead focuses on enhancing models' local performance under data heterogeneity via model personalization techniques, such as partial parameter sharing \cite{liang2020think, sun2021partialfed, husnoo2022fedrep}, meta-learning\cite{li2021ditto, fallah2020personalized}, latent representation space alignment\cite{tan2022fedproto,oh2021fedbabu},  hyper-network\cite{shamsian2021personalized}, and knowledge decoupling\cite{chen2021bridging,zhang2023gpfl}. Nevertheless, most of these methods are naturally designed for and verified in label skew scenarios \cite{kairouz2021advances}, overlooking the domain shift problem and resulting in suboptimal performance in cross-domain scenarios \cite{chen2024fair}.

% To address data heterogeneity, previous works 
% a pioneering work is FedProx (Li et al.), which introduced a proximal term to the local optimization objective. This approach encourages updates to remain closer to the global model, thereby mitigating the impact of divergent data distributions. Building on this, FedDyn (Acar et al.) proposed a dynamic regularization method. This method adjusts the regularization strength based on local gradient dissimilarity, enhancing convergence in situations with label skew. Besides, Scaffold (Karimireddy et al.) took a different approach by employing control variates to correct for client drift caused by heterogeneous data. This method reduces the variance introduced by disparate label distributions, thus improving the overall performance of the federated model. Moon (Li et al.) introduced a momentum-based correction term that dynamically adjusts during the training process to counteract the effects of label heterogeneity, promoting faster and more stable convergence. Moreover, FedSpeed (Xu et al.) leveraged local stochastic gradient descent (SGD) with momentum and adaptive learning rates to accelerate convergence while effectively handling label distribution shifts across clients. However, most existing works are primarily designed for label skew scenarios, leading to suboptimal results in feature skew scenarios. \textcolor{red}{In this paper, we introduce}

\subsection{Cross-domain FL}
Cross-domain FL refers to FL with clients whose datasets come from different domains \cite{chen2024fair, sun2021partialfed, huang2023rethinking, zhou2023fedfa, li2021fedbn, liu2021feddg, peng2019federated, jiang2022harmofl}. The domain shift across clients' datasets can hinder the FL model training process, leading to model performance degradation \cite{chen2024fair}. Previous methods addressing this issue can be mainly grouped into two sets. Methods in the first set focus on maximizing different clients' consensus across domains at different levels. For example, \cite{jiang2022harmofl} exchanges a few local data information across clients to normalize their samples in the frequency space at the input level. \cite{zhou2023fedfa} increases the visibility of all domains' samples to each client during local training by feature augmentation in the intermediate layers of the model, leading to improved global consensus at the feature level. \cite{tan2022fedproto, huang2023rethinking} maintain consistent class prototypes before the output layers of the model to increase clients' consensus in the representation space. \cite{chen2024fair} instead enhances clients' consensus directly at the model parameter level by minimizing the conflicts of model updates of important parameters. Another set of methods uses model personalization techniques to help suit the global model to their local domain skew. \cite{li2021fedbn} keeps batch normalization layers to be locally maintained to make the model more adaptable to their personal domains. \cite{sun2021partialfed} adaptively loads partial parameters instead of the entire model for each local training. Despite the rapid development of the two sets of methods respectively, the way to integrate the advantages of the two sets still remains unexplored, leading to a potentially large space for further performance improvement.

To this end, we rethink the domain shift problem in FL from a hybrid view of existing methods (e.g., consensus enhancement and personalization). Different from previous studies, we fine-grained identify what should be personalized (e.g., domain-specific skew erasing) and on which the consensus should be maximized (the treatment to deskewed features), thus successfully bridging the two distinct insights behind existing approaches. 
\begin{figure*}[t]
    \centering
    \includegraphics[width=\linewidth]{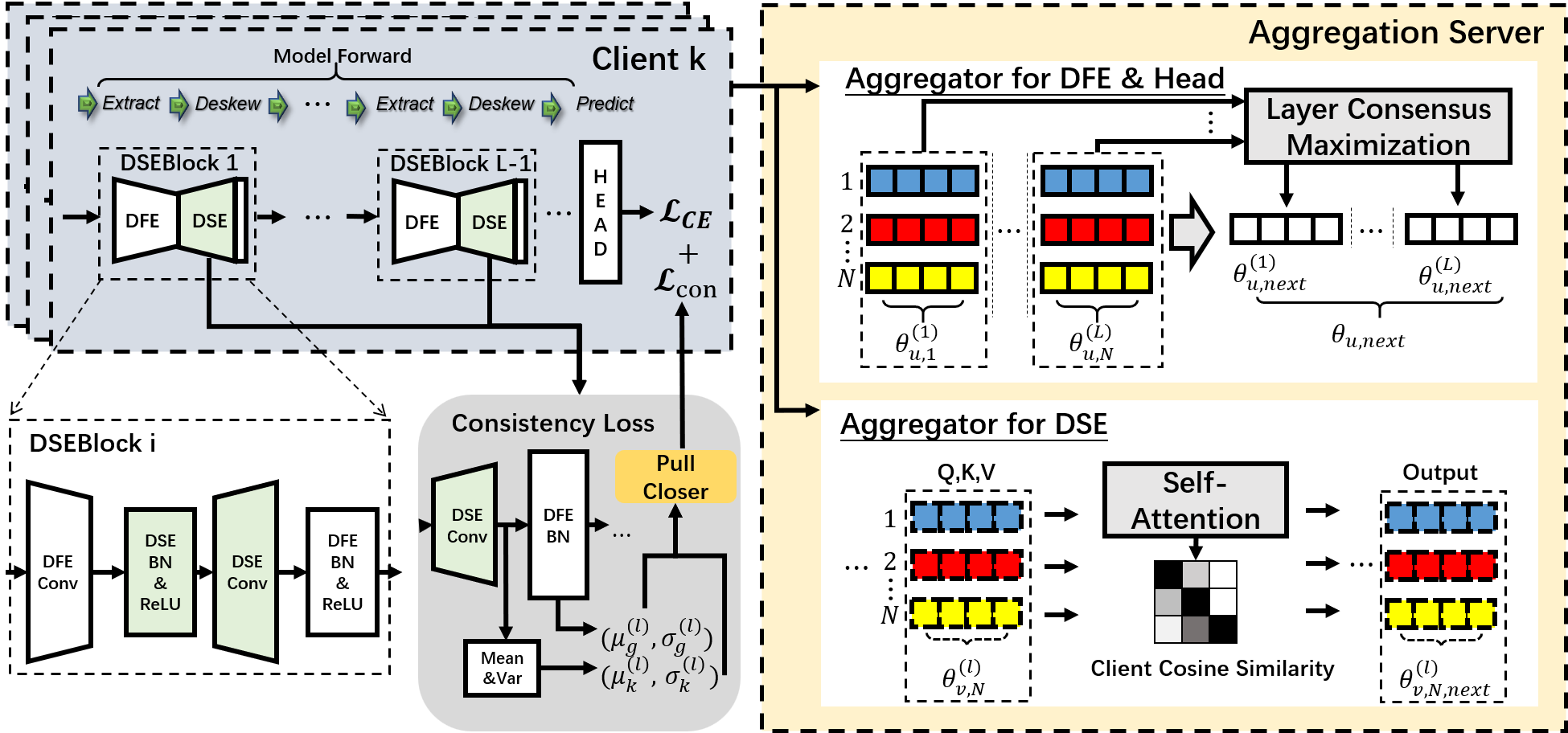}
    \caption{The overview of the FDSE framework.}  
    \label{fig2}
    \vspace{-0.5cm}
\end{figure*}
\section{Problem Formulation}
\paragraph{Personalized FL.}  In personalized FL, there exist $N$ clients equipped with their private data $\{\mathcal{D}_i| i\in [N]\}$, and the goal is to obtain a series of model $\mathbf{\Theta} = \{\mathbf{\theta}_1,\mathbf{\theta}_2, ..., \mathbf{\theta}_N\}$ that minimize
 \begin{equation}
     \min_{\mathbf{\Theta}}=\sum_{i\in [N]}\frac{|\mathcal{D}_i|}{|\mathcal{D}|} F_i, F_i={\mathbb{E}_{(x,y)\sim \mathcal{D}_i}\left[\ell( \mathbf{\theta}_i; x,y)\right]}  
 \end{equation}
 where  $|\mathcal{D}|=\sum_{i=1}^N\mathcal{D}_i$ and $\ell(\cdot)$ is  the loss function. In this paper, we consider one popular scheme of personalized FL where some parameters are globally shared among all the clients and the others are privately kept \cite{husnoo2022fedrep,barro1992convergence}. Let $\mathbf{\theta}_i=[\mathbf{\theta}_{u};\mathbf{\theta}_{v,i}]$, we denote $\mathbf{\theta}_u$ the globally shared parameters and $\mathbf{\theta}_{v,i}$ the personalized ones.
% Federated Learning. Following typical Federated Learning setup [29, 36, 37], we consider there are M clients (indexed by m). Each client holds private data Dm =  {xi, yi}Nm  i=1, where Nm represents the data size of client m. The optimization objective of FL is to minimize global loss:  mwin F (w) =  M  X  m=1  pmfm(w),(1)  where fm(w) = 1  Nm  PNm  i=1 L(xi, yi; w), pm is the weight of client. L(xi, yi; w) is the loss of data (xi, yi) with model parameters w. Each client updates its model locally, and the server then aggregates model updates from all clients.
\paragraph{Cross-domain FL}  Cross-domain FL \cite{chen2024fair, zhou2023fedfa} refers to FL with clients that exhibit domain shift among their local datasets. The domain shift means that different clients share the consistent label space $\mathbb{P}(Y)$ but differ in conditional feature distribution $\mathbb{P}(X|Y)$ given $Y$ (e.g., $\mathbb{P}_i(X|Y=y)\ne \mathbb{P}_j(X|Y=y), \forall i\ne j, i,j\in[N]$). 
% You must include your signed IEEE copyright release form when you submit your finished paper.
% We MUST have this form before your paper can be published in the proceedings.

% Please direct any questions to the production editor in charge of these proceedings at the IEEE Computer Society Press:
% \url{https://www.computer.org/about/contact}.

\section{Methodology}
In this section, we present the main procedure and implementation of FDSE. We begin by illustrating the layer decomposition approach, which allows us to decouple the personalized domain skew erasing from the common task objective by assigning them to different model parameters in Sec. \ref{sec_decom}. Next, we introduce the consistency regularization used during clients' local training to enhance the effectiveness of feature deskewing in Sec. \ref{sec_reg}. Finally, we discuss the model aggregation strategies respectively for common and personalized parameters in Sec. \ref{sec_agg}.

\subsection{Layer Decomposition}\label{sec_decom}
As aforementioned in Sec.\ref{sec:intro}, we personalize partial model parameters for each client to erase their domain-specific skew. Existing deskewing methods usually align cross-domain features at a certain layer of the neural network. For example, \cite{jiang2022harmofl, yuan2024density} align samples' attributions in the feature space before feeding them into the model, and \cite{tan2022fedproto, liang2020think} mitigate the domain divergence in the representation space at the output layer.
However, premature deskewing  (e.g., the input layer) may lead to insufficient skew elimination while the too-delayed one (e.g., the output layer ) can hinder domain-agnostic knowledge extraction. Therefore, instead of only deskewing features at a certain layer, we view the model forward passing as an iterative deskewing process that extracts and then deskews features alternatively. 
This is efficiently achieved by decomposing each original layer in the neural network into a Domain-agnostic Feature Extractor (DFE) and a Domain-specific Skew Eraser (DSE), as is shown in Figure \ref{fig2}. 
% we aim to gradually deskew features layer-by-layer during the model forward passing. 
Concretely, given a convolution layer $f$ with filter size $k$, input channel number $S$, and output channel number $T$, we follow \cite{han2020ghostnet} to decompose it into two sub-convolution modules $f_{\text{DSE}}\circ f_{\text{DFE}}$ where each submodule consists of a convolution, a batch normalization layer, and an activation function. The DFE module shares parameters with the original layer $f$ (e.g., kernel size, stride, and padding) except for the number of output channels $T_{\text{DFE}}=\lceil T/G \rceil$  by
\begin{equation}
    \bold X_{\text{DFE}} = f_{\text{DFE}}*\bold X
\end{equation}
where $\bold X_{\text{DFE}} \in \mathbb{R}^{T_{\text{DFE}}\times h\times w}$, $h,w$ are respectively the height and the weight of the original output $f(\bold X)$ and $G$ is the architecture parameter. Then, the DFE module maps each channel in the DFE module's output into $G$ channels via cheap linear operations by
\begin{align}
\bold X_{\text{DSE},i} &=f_{\text{DSE},i}*\text{ReLU}\left(\text{BN}_{\text{DSE}}\left(\bold X_{\text{DFE},i}\right)\right), i\in[T_{\text{DFE}}]\notag\\
    &\bold X_{\text{out}}=\text{ReLU}\left(\text{BN}_{\text{DFE}}\left(\text{Concat}(\{\bold X_{\text{DSE},i}\})\right)\right)
\end{align}
where $\bold X_{\text{DFE},i} \in \mathbb{R}^{1\times h\times w},\bold X_{\text{DSE},i} \in \mathbb{R}^{G\times h\times w},\bold X_{\text{out}} \in \mathbb{R}^{T\times h\times w}$, and each $f_{\text{DSE},i}$ augments each channel $\bold X_{\text{DFE},i}$ to $G$ channels. 
Although there have been plenty of works that decompose convolution layers for efficiency \cite{tan2019efficientnet, tan2021efficientnetv2, mei2022flanc}, this layer separation has the natural advantage for our objective decoupling purpose. On one hand, 
the core components of the output features are independently extracted by the DFE module, each of which is then cheaply expanded to several similar variants by the DSE module. This paradigm enables the DFE module to learn the primary knowledge (e.g., the core components of feature maps), making it suitable for modeling domain-agnostic information. On the other hand, the simplicity of the DSE module avoids over-transforming features during deskewing processes, e.g., $94\times$ smaller than the DFE module in a convolution with 64 channels and kernel size 5. We further enhance the knowledge decoupling for the two modules via regularization term in Sec.\ref{sec_reg} and the parameter sharing strategy in Sec.\ref{sec_agg}.

\subsection{Consistency Regularization}\label{sec_reg}
To make the DSE module focus on erasing domain-specific skew for each client, we consider regularizing the DSE module's output features during clients' local training. Supposing that the $l$th DSE module $f_{\text{DSE}}^{(k,l)}$ in the model has ideally erased the domain skew for the $k$th client's local data, then the BN layer of the next DFE module $\text{BN}_{\text{DFE}}^{(k,l)}$ cannot infer the samples' domain via the distribution of the seen data when there is no label skew. Based on this insight, we manually pull the statistics of DSE's output to be close to the global statistics of the corresponding DFE module, thus enhancing the deskewing characteristic of the DSE module from the statistical view. 
% We assume that the local feature $\bold X_{k}^{(l)}$ of the $l$th layer follows the gaussian distribution $\mathcal{N}\left(\mu_k^{(l)}, \text{Diag}(\sigma_k^{(l)2})\right)$ and the globally consistent feature follows $\mathcal{N}\left(\mu_g^{(l)}, \text{Diag}(\sigma_g^{(l)2})\right)$. This assumption has been widely made in previous works. Then, we minimize the KL-divergence between the two distributions to pull the skewed local distribution closer to the globally consistent one by
% \begin{equation}
%     D_{KL}\left(\mathcal{N}\left(\mu_g^{(l)}, \text{Diag}(\sigma_g^{(l)2})\right)||\mathcal{N}\left(\mu_k^{(l)}, \text{Diag}(\sigma_k^{(l)2})\right)\right)
% \end{equation}
Specifically, given the $b$th batch feature $\bold X_{k,b}^{(l)}, |\bold X_{k,b}^{(l)}|=B$ fed to the $l$th layer of the model, we first compute its statistics (i.e., mean and variance) by
\begin{equation}
\mathbf{\mu}_{k,b}^{(l)}=\frac{1}{B}\sum_{i=1}^B \bold X_{k,b,i}^{(l)}, \mathbf{\sigma}_{k,b}^{(l)2}=\frac{1}{B}\sum_{i=1}^B \left(\bold X_{k,b,i}^{(l)}-\mathbf{\mu}_{k,b}^{(l)}\right)^2
\end{equation}
Then, we estimate the local running statistics by exponential averaging with the momentum coefficient $\gamma$ of $\text{BN}_{\text{DFE}}$ during local training
\begin{align}
\hat\mu_{k,b}^{(l)}=\gamma\hat\mu_{k,b-1}^{(l)}+(1-\gamma)\mu_{k,b}^{(l)}\notag\\
\hat\sigma_{k,b}^{(l)2}=\gamma\hat\sigma_{k,b-1}^{(l)2}+(1-\gamma)\sigma_{k,b}^{(l)2}
\end{align}
The local statistics of the $l$th layer are finally pulled close to the global ones by
\begin{equation}
\mathcal{L}^{(l)}_{\text{Con}} = \frac{1}{d}\left\|\hat{\mu}^{(l)}_{k,b} - \mu^{(l)}_{g}\right\|^2 + \left(\frac{\|\hat{\sigma}^{(l)2}_{k,b}\|_1 - \|\sigma^{(l)2}_{g}\|_1}{d}\right)^2
\end{equation}
% \begin{align}
%     \mathcal{L}_{\text{Con}}^{(l)} = \frac{1}{d}\sum_{i=1}^d\left(\log{\frac{\hat\sigma_{k,b,i}^{(l)}}{\sigma_{g,i}^{(l)}}}+\frac{\hat\sigma_{g,i}^{(l)2}+(\mu_{g,i}^{(l)2}-\hat\mu_{k,b,i}^{(l)2})}{2\hat\sigma_{k,b,i}^{(l)2}}\right)
% \end{align}
where $\mu_{g}^{(l)}, \sigma_{g}^{(l)}$ are the corresponding layer's global statistics in the received global model and $d$ is the feature dimension. In this way, both the centers and the sizes of the clients' feature spaces are regularized to be consistent. This rule is then applied to each DSE module in a depth-increasing manner, as formulated in the consistency regularization term
\begin{equation}
     \mathcal{L}_{\text{Con}}=\sum_{l=1}^L w_{l}\mathcal{L}_{\text{Con}}^{(l)}, \text{w.r.t., }w_{l}=\frac{\exp{(\beta l)}}{\sum_{l=1}^L\exp{(\beta l)}}
\end{equation}
where $L$ is the number of DSE modules and $\beta$ is the hyper-parameter that enables gradual features deskewing across layers. A smaller $\beta$ can strengthen feature deskewing by emphasizing each layer's statistical consistency. We fix $\beta=0.001$ in practice and only tune the coefficient $\lambda$ that scales the regularization term to balance the task objective and the regularization. In addition, this consistency regularization can help adapt the trained model to new unseen domains without any labels, where we fine-tune the DSE modules to minimize the regularization loss as depicted in Sec. \ref{sec_unseen}.

\subsection{Model Aggregation}\label{sec_agg}
We use different strategies to aggregate the two fundamental components of FDSE (e.g., DFE and DSE). For the DFE modules, we share them among all the clients and aggregate them through a fair layer consensus maximization mechanism. For the DSE modules, we personalized them for each client based on similarity-aware aggregation to erase heterogeneous domain skew collaboratively.
As a result, the varying degrees of parameter sharing further facilitate the distinct knowledge acquisition of the DFE and DSE modules in a decoupled manner, as the local knowledge encoded in the DFE modules is much more frequently smoothed over all the clients than the DSE modules. We illustrate the details of the two aggregation strategies below.

% The difference in parameter-sharing degrees further promotes the knowledge to be respectively learned by the DFE and DSE modules in a decoupled way, since local knowledge encoded in the DFE modules will be continuously smoothed among different clients while the local knowledge encoded in the DSE modules has a much lower chance to be mixed with others. 

% we implement the proposed domain-agnostic fair treatment via the . For the personalized parameters (e.g., DSE modules), we differentially aggregate them to collaboratively learn domain-specific skew elimination for similar clients. We visualize the two aggregation processes in the right block of Figure \ref{fig2}.
\subsubsection{Consensus Maximization}\label{sec_conagg} 
We remark that DFE modules are designed to model domain-agnostic knowledge. However, model updates from different clients may largely conflict with each other due to severe domain shifts \cite{wang2021fedfv, chen2024fair, hu2022federated}, preventing the model from learning domain-agnostic knowledge by biasing the training process towards the dominant clients. 
% Despite the nature of feature deskewing of FDSE, the conflicts in model updates can still degrade model performance due to the gradual domain skew erasing across layers. 
To this end, we propose to maximize client consensus when aggregating fully shared parameters (e.g., DFE modules and the task head) to enhance domain-agnostic knowledge learning. Motivated by the multi-gradient-descent-algorihtm \cite{hu2022federated,desideri2009multiple} that optimizes the model update to be harmonies with each client's update,  we minimize the L2-norm of the aggregated update for each layer to maximize layer-wise consensus. Specifically, given clients' local updates on DSE modules and heads $\{\Delta{\theta}_{u,k}^{(l)}\}, k\in[N]$, we aggregate them by 
\begin{align}
\bar d_{u}^{(l)}&=\frac{1}{N}\sum_{k=1}^N\|\Delta{\theta}_{u,k}^{(l)}\|_2,\bold d_{u,k}^{(l)} = \frac{\Delta{\theta}_{u,k}^{(l)}}{\|\Delta{\theta}_{u,k}^{(l)}\|_2}\notag\\
\Delta{\theta}_{u}^{(l)}&=\bar d_{u}^{(l)}\sum_{k=1}^Nu_{k}^{(l)} \bold d_{u,k}^{(l)},\bold u^{(l)} = \text{argmin}_{\bold u}\|\sum_{k=1}^Nu_{k} \bold d_{u,k}^{(l)}\|_2^2\label{eq_sagg}
\end{align}
where $\Delta{\theta}_{u}^{(l)}\cdot \Delta{\theta}_{u,k}^{(l)}\ge 0, \forall k$ is guaranteed by the above optimization objective \cite{desideri2009multiple}. This aggregation scheme enhances the consensus in model updates across clients without sacrificing anyone's benefit. Besides, this process is independently repeated for different layers since simply mitigating conflicts at the model level cannot prevent the final update from favoring parts of clients at some layers, reducing the layer utility for those clients \cite{pan2024fedlf}.
\begin{table*}\small
\centering
\begin{tblr}{
  cells = {c},
  cell{1}{1} = {r=2}{},
  cell{1}{2} = {c=2}{},
  cell{1}{4} = {c=2}{},
  cell{1}{6} = {c=2}{},
  vline{2} = {-}{},
  hline{1,3,15, 16} = {-}{},
 hline{2} = {2-7}{},
}
\hline
\textbf{Method} & \textbf{Domainnet} &              & \textbf{Office-Caltech10} &              & \textbf{PACS} &              \\
                & \textbf{ALL}       & \textbf{AVG} & \textbf{ALL}              & \textbf{AVG} & \textbf{ALL}  & \textbf{AVG} \\
\textbf{Local} &$57.10_{\pm0.32}$&$52.96_{\pm0.33}$&$64.47_{\pm2.52}$&$62.72_{\pm7.81}$&$61.29_{\pm2.47}$&$57.16_{\pm2.85}$\\
\textbf{FedAvg} (AISTATS 2017) &$69.17_{\pm0.46}$&$67.53_{\pm0.41}$&$82.60_{\pm3.14}$&$86.26_{\pm2.54}$&$74.30_{\pm1.90}$&$72.10_{\pm1.42}$\\
\textbf{LG-FedAvg} (NIPSW 2019) &$71.13_{\pm0.30}$&$67.97_{\pm0.32}$&$82.69_{\pm0.53}$&$87.29_{\pm1.32}$&$79.11_{\pm0.69}$&$76.72_{\pm0.54}$\\
\textbf{FedProx} (MLSys 2020) &$68.81_{\pm0.71}$&$67.47_{\pm0.66}$&$82.69_{\pm1.52}$&$87.36_{\pm1.87}$&$74.38_{\pm1.55}$&$72.33_{\pm1.53}$\\
\textbf{Scaffold} (ICML 2020)&$70.41_{\pm0.40}$&$69.06_{\pm0.43}$&$80.86_{\pm2.43}$&$85.87_{\pm1.87}$&$76.30_{\pm0.93}$&$74.26_{\pm0.95}$\\
\textbf{FedDyn} (ICLR 2021)&$70.02_{\pm0.57}$&$68.86_{\pm0.54}$&$82.77_{\pm1.82}$&$87.80_{\pm1.96}$&$74.92_{\pm1.29}$&$73.19_{\pm1.01}$\\
\textbf{MOON} (CVPR 2021) &$68.35_{\pm0.32}$&$66.65_{\pm0.29}$&$80.12_{\pm1.86}$&$82.48_{\pm1.71}$&$75.00_{\pm0.32}$&$72.13_{\pm0.32}$\\
\textbf{Ditto} (ICML 2021)&$75.18_{\pm0.37}$&$72.82_{\pm0.35}$&$84.12_{\pm1.32}$&$88.72_{\pm1.28}$&$82.02_{\pm1.32}$&$80.03_{\pm1.37}$\\
\textbf{PartialFed} (NIPS 2021)&$75.92_{\pm0.24}$&$73.46_{\pm0.41}$&$82.70_{\pm2.12}$&$88.33_{\pm2.17}$&$81.22_{\pm0.98}$&$79.18_{\pm1.09}$\\
\textbf{FedBN} (ICLR 2021)&$74.75_{\pm0.24}$&$72.25_{\pm0.20}$&$83.08_{\pm1.84}$&$87.01_{\pm1.30}$&$81.58_{\pm0.79}$&$79.47_{\pm0.69}$\\
\textbf{FedFA} (ICLR 2023) &$69.47_{\pm0.29}$&$67.60_{\pm0.29}$&$82.98_{\pm2.84}$&$86.69_{\pm3.02}$&$75.44_{\pm0.71}$&$73.60_{\pm0.86}$\\
\textbf{FedHeal} (CVPR 2024) &$69.43_{\pm0.71}$&$67.96_{\pm0.65}$&$81.73_{\pm3.19}$&$86.29_{\pm2.71}$&$75.46_{\pm0.84}$&$73.51_{\pm0.83}$\\
\textbf{FDSE (Ours)} &$\bold{76.77_{\pm0.41}}$&$\bold{74.50_{\pm0.40}}$&$\bold{87.15_{\pm2.06}}$&$\bold{91.58_{\pm2.01}}$&$\bold{83.81_{\pm1.70}}$&$\bold{82.17_{\pm1.49}}$\\
\end{tblr}
\caption{Comparison of model testing accuracy (\%)$\uparrow$ on DomainNet, Office-Caltech10, and PACS datasets. The optimal results are marked by \textbf{bold}. \textbf{ALL} refers to the model testing accuracy on all clients' local testing samples and \textbf{AVG} refers to the mean of of clients' local testing accuracies. Each result is averaged over 5 trials with different fixed random seeds.}\label{table_main}
\vspace{-0.2cm}
\end{table*}

\subsubsection{Similarity-aware Personalization}\label{sec_simagg} 
To enable collaboratively erasing domain skew for clients from similar domains, we aggregate each DSE module's parameters with a self-attention module on the server side based on clients' similarity. Given clients's parameters of the $l$th DSE module $\{\mathbf{\theta}_{v,k}^{(l)},k\in[N]\}$, we aggregate them by 
\begin{align}
     &\bold Q_l =  \bold K_l=\left[\frac{\bold V_{lk}}{\|\bold V_{lk}\|_2}\right]^\top|_{k=1\text{ to }N},  \bold V_l = \left[\mathbf{\theta}_{v,1}^{(l)}, \cdots , \mathbf{\theta}_{v,N}^{(l)}\right]^\top  \notag\\ 
    &\left[\mathbf{\theta}_{v,1, \text{next}}^{(l)}, \cdots , \mathbf{\theta}_{v,N, \text{next}}^{(l)}\right]^\top =\text{softmax}\left(\frac{\bold Q_l \bold K_l^\top}{\tau}\right)\bold V_l\label{eq_pagg}
\end{align}
where $\tau$ is the temperature parameter controlling the degree of personalization of DSE modules (e.g., smaller $\tau$ corresponds to a higher personalization degree and stronger skew elimination). We also perform the personalized aggregation independently for each layer's DSE module, since the optimal personalization degree may vary across layers \cite{ma2022layer}. For the non-trainable statistical parameters, we did not aggregate them for $\text{BN}_{\text{DSE}}$ and directly average them for $\text{BN}_{\text{DFE}}$.
The pseudo-code of FDSE is summarized in Algorithm 1.

\begin{algorithm}[tb]
    \caption{Federated Domain Shift Eraser}
    \label{alg:1}
    \textbf{Input}:The global model $\mathcal{M}$, the number of local epochs $E$, and the learning rate $\eta_t$\\
    \begin{algorithmic}[1]
    \STATE Decompose the initial model into globally shared parameters $\theta_u^{0}$ and personalized parameters $\theta_v^{0}$ in Sec. \ref{sec_decom}
    \STATE Initialize clients' personalized parameters $\theta_{v,k}^0=\theta_v^{0}, \forall k\in[N]$
    \FOR{ communication round $t = 0,1,...,T-1$}
    % \STATE The server checks the set of available clients $A_t$.
    % \STATE The server uses $\bold v^t$ and $\bold H$ to solve equation (16) within the maximum wall-clock time $\tau_{max}$ to obtain the sampled client set $S_t\subseteq A_t$
    \STATE The server broadcasts the model $\theta_{k}^t=(\theta^t_u,\theta^t_{v,k} )$ to each client $k$.
    \FOR{ each client $k\in [N]$}
    \FOR{ each iteration $i=0,1,...,E-1$}
    \STATE Compute loss $\mathcal{L}_k=\mathcal{L}_{\text{task}}+\lambda\mathcal{L}_{\text{Con}}$
    \STATE $\theta_{k,i+1}^{t}\leftarrow \theta_{k,i}^{t}-\eta_t\nabla \mathcal{L}_k(\theta_{k,i}^{t})$ 
    \ENDFOR
    \STATE Client $k$ send the model parameters $\theta_{k}^{t+1}=\theta_{k,E}^{t}$ to the server.
    \ENDFOR
    \STATE The server respectively aggregates the received globally shared model parameters $    \theta^{t+1}_u $ by Eq. (\ref{eq_sagg}) and personalized model parameters $    \left[\theta^{t+1}_{v,k}\right]^\top $ by Eq. (\ref{eq_pagg})
    % \STATE The server updates t he    sampling counts of clients $\bold v^{t+1}[k]\leftarrow \bold v^{t}[k] + \mathbb{I}(k\in S_t)$
    \ENDFOR
    \end{algorithmic}
    \end{algorithm}
    
% W = V · U with V ∈ Rk2×S×R, U ∈ RR×T , and R ≤ T . The factorization for fully-connected layer is equivalent as the k = 1 case. Such low-rank approximation is effective for model compression in order to reduce model size and computation cost [30, 31, 32, 33]. Similarly, we speculate federated learning with system heterogeneity could benefit from tensor decomposition.  Thus, we introduce All-In-One Neural Composition, that employs low-rank approximation to represent networks in different widths with a unified expression. For an arbitrary p-width network, we decouple its weight Wp ∈ Rk2×Sp×T p as a shared tensor Vshare and capacity-specific tensor Up, i.e.,

\section{Experiments}\label{sec:exp}
\subsection{Setup}
\paragraph{Dataset \& Model.}  We evaluate our method on three popular multi-domain image classification tasks: Office-Caltech10 \cite{sun2021partialfed}, DomainNet \cite{leventidis2023domainnet}, and PACS \cite{zhou2020pacs}. We follow \cite{zhou2023fedfa,li2021fedbn} to use AlexNet as the backbone for these datasets and allocate a single domain's data to each client respectively for all three datasets. 
\paragraph{Baselines.} We compare our method with four types of baselines: 1) Local training only; 2) Vallina FL with data heterogeneity: FedAvg \cite{mcmahan2017communication}, FedProx \cite{li2020federated}, Scaffold \cite{karimireddy2020scaffold}, FedDyn \cite{acar2021federated}, MOON \cite{li2021model} 3) Consensus-based FL: FedHeal \cite{chen2024fair}, FedFA \cite{zhou2023fedfa}; 4) Personalized FL:  LG-FedAvg \cite{liang2020think}, FedBN \cite{li2021fedbn}, PartialFed \cite{sun2021partialfed}.

\paragraph{Hyper-parameters.} We tune the learning rate $\eta\in\{0.001, 0.01, 0.05, 0.1, 0.5\}$ by grid search for each method. The batch size is fixed to $50$ and the local epochs for Domainnet, Office-Caltech10, and PACS are respectively $5$, $1$, and $5$. We run each trial for 500 communication rounds with the learning rate decay ratio $0.998$ per round. All the methods' algorithmic hyper-parameters are respectively tuned to their optimal. More details on the hyper-parameter setting are in the supplementary materials.

\subsubsection{Implementation} All our experiments are run on a 64 GB-RAM Ubuntu 22.04.3 server with Intel(R) Xeon(R) CPU E5-2630 v4 @ 2.20GHz and 2 NVidia(R) RTX4090 GPUs. All code is implemented in PyTorch 1.12.0 and FLGo 0.3.29 \cite{wang2023flgo}. 
\begin{figure}
    \centering
    \includegraphics[width=\linewidth]{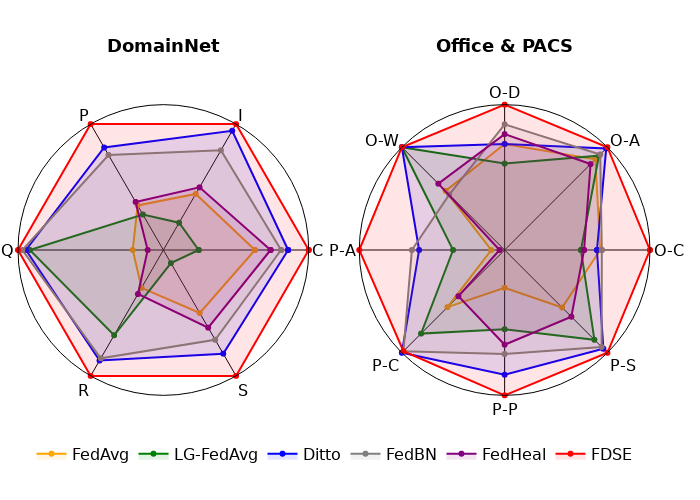}
    \caption{Evaluation results on individual domains (i.e., clients) across the three datasets. Each axis represents the result for a specific domain and is scaled by the axis's maximum value for clarity.}
    \label{fig3_polar}
    \vspace{-0.5cm}
\end{figure}

\subsection{Comparison with Baselines}
\begin{figure*}
    \centering
    \includegraphics[width=0.8\linewidth]{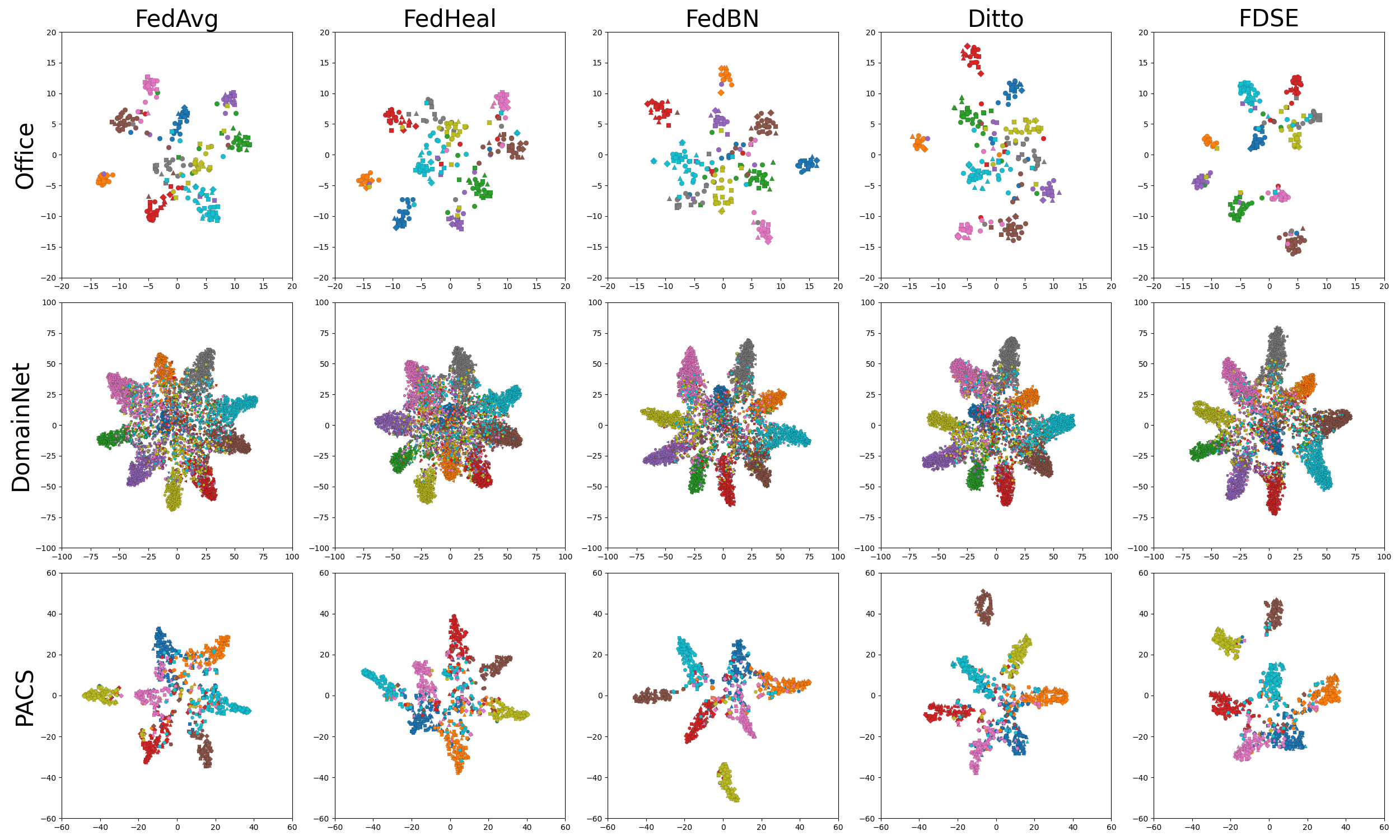}
    \caption{T-SNE visualization for representation space of different methods on Office-Caltech10, DomainNet, and PACS. Each color represents one class of samples and each shape represents one domain.}
    \label{fig_tsne}
\end{figure*}
\begin{figure*}
    \centering
    \includegraphics[width=0.8\linewidth]{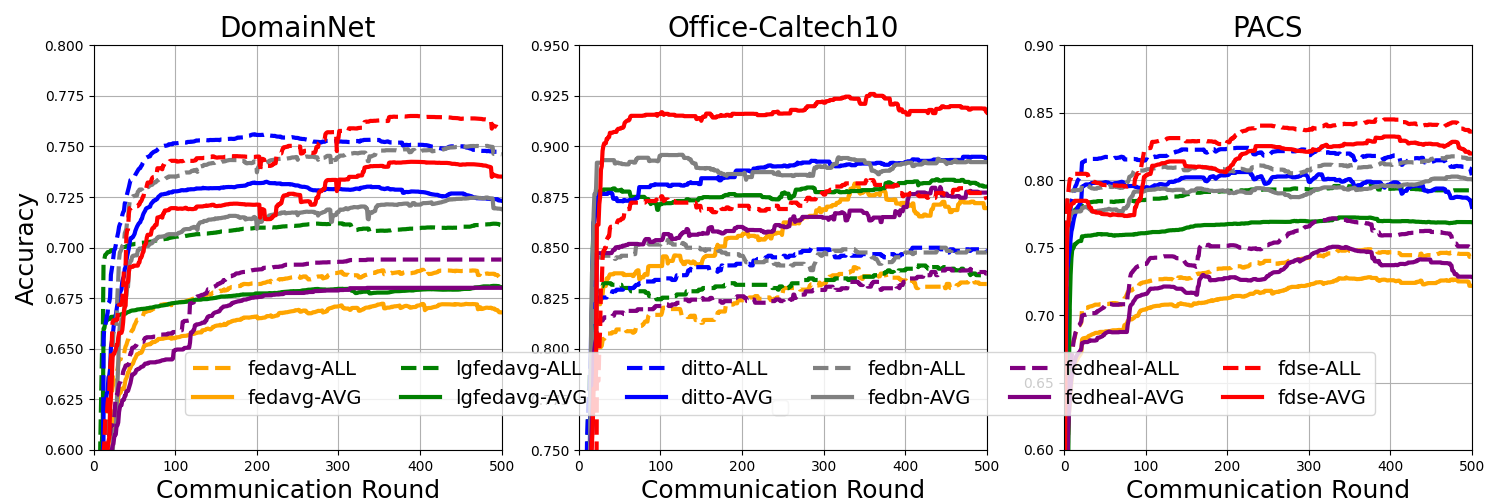}
    \caption{Testing accuracy v.s. communication rounds.}\label{fig_curve}
    \vspace{-0.2cm}
\end{figure*}

\paragraph{Overall Performance.} Table \ref{table_main} compares our proposed FDSE with several baselines. Notably, FDSE consistently achieves optimal results across all settings, demonstrating a significant advantage in enhancing model performance in cross-domain FL.
Additionally, we observe that personalized methods (e.g., LG-FedAvg, Ditto, and FedBN) outperform almost all non-personalized approaches (e.g., FedAvg, FedHeal, and FedFA). This underscores the importance of incorporating domain-specific knowledge for each domain. Furthermore, enhancing client consensus also contributes to performance improvement, as evidenced by FedHeal and FedFA consistently outperforming FedAvg in nearly all settings. Thus, we attribute the superiority of FDSE to the success in integrating the complementary advantages of both consense-based and personalization-based methods.
\paragraph{Individual Performance.} Figure \ref{fig3_polar} illustrates the individual performance of clients for different methods, where a larger area under a method's curve indicates better performance. FDSE outperforms all other methods across nearly all clients, confirming its effectiveness across all domains rather than just specific ones.
\paragraph{Convergence.} We plot the testing accuracy across communication rounds in Figure \ref{fig_curve}. While Ditto demonstrates a slightly faster convergence speed than FDSE in the early stages of training on DomainNet and PACS, FDSE ultimately achieves higher performance as training progresses.
% We highlight that FDSE achieves optimal results across all the settings, indicating the superior advantage in enhancing model performance in cross-domain FL against other baselines.
% Besides, we notice that the personalized methods (e.g., LG-FedAvg, Ditto, and FedBN) dominate most of the non-personalized ones (e.g., FedAvg, FedHeal, and FedFA) under both two metrics, revealing the importance of modeling domain-specific knowledge for each domain. On the other hand, enhancing clients' consensus can also achieve better model performance (e.g., FedHeal outperforms FedAvg over almost all the settings). 

% By integrating the advantages of both types of methods, our proposed method (FDSE) consistently achieves optimal results across all the scenarios, e.g., at most $+10.07\%$ against FedAvg in PACS. We visualize each client's local testing accuracy in Figure \ref{fig3_polar}. Our FDSE achieves the optimal local model performance for all the clients across different datasets. We finally show the learning curve (e.g., testing accuracy v.s. communication rounds) in Figure \ref{fig_curve}, where our methods exhibit faster convergence and higher performance than other baselines. These results confirm the superior advantage of our FDSE in cross-domain FL.
% ,  

\begin{table*}
\centering
\begin{tabular}{c|ccccc|ccccccc} 
\hline\hline
\multirow{2}{*}{\textbf{Method}} & \multicolumn{5}{c|}{\textbf{Office-Caltech10}}                                     & \multicolumn{7}{c}{\textbf{\textbf{DomainNet}}}                                                                                                            \\
                                 & \textbf{C}     & \textbf{A}     & \textbf{D}     & \textbf{W}     & \textbf{AVG}   & \textbf{\textbf{C}} & \textbf{\textbf{I}} & \textbf{\textbf{P}} & \textbf{\textbf{Q}} & \textbf{\textbf{R}} & \textbf{\textbf{S}} & \textbf{\textbf{AVG}}  \\ 
\hline
\textbf{FedAvg}                  & $51.78$        & $70.52$        & $80.00$        & $65.51$        & $66.95$        & $62.81$             & $30.15$             & $55.53$             & $48.86$             & $59.74$             & $58.92$             & $52.66$                \\
\textbf{FedBN}                   & $\bold{60.71}$ & $70.52$        & $80.00$        & $55.17$        & $66.60$        & $62.56$             & $31.14$             & $57.26$             & $53.06$             & $63.17$             & $62.46$             & $54.94$                \\
\textbf{- align}                 & $40.17$        & $68.42$        & $86.66$        & $72.41$        & $66.91 $       & $60.53$             & $31.57$             & $54.33$             & $50.60$             & $59.85$             & $62.59$             & $53.24$                \\
\textbf{FDSE}                    & $57.14$        & $\bold{75.78}$ & $\bold{93.33}$ & $\bold{75.86}$ & $\bold{75.52}$ & $\bold{65.22}$      & $\bold{32.34}$      & $\bold{59.32}$      & $\bold{55.00}$      & $\bold{64.28}$      & $\bold{65.28}$      & $\bold{56.91}$         \\
\cmidrule{1-13}
\end{tabular}
\caption{Model performance ($\uparrow$) of adapting the trained model to different unseen clients. The optimal results are marked by \textbf{bold}.}\label{table_unseen}
\vspace{-0.5cm}
\end{table*}
\subsection{T-SNE Visualization}\label{sec_tsne}
Figure \ref{fig_tsne} presents the t-SNE visualization analysis of the representation space (i.e., the feature space before the output layer of the model) for different methods. On one hand, FDSE increases the inter-class distances of samples, with a greater separation between samples of different colors compared to other baselines. On the other hand, FDSE reduces the intra-class distance among samples of the same color, resulting in tighter color clusters than those observed in other baselines. This confirms FDSE’s ability to enhance model performance while reducing domain skew.

% Figure \ref{fig_tsne} shows the t-SNE visualization analysis on the representation space (i.e., feature space before the output layer of the model) for different methods. On one hand, FDSE increases the inter-class distances, where the gap between samples with different colors are larger than other baselines.  On the other hand, FDSE decreases the intra-class distance between samples with the same color, where the area of each color cluster of FDSE is smaller than other baselines. This confirms the FDSE's ability to both enhance model performance and diminish domain skew.

% Compared with non-personalized methods, our FDSE achieves larger distances between different class centers. Compared with personalized methods, FDSE learns tighter clusters for each label in the representation space regardless of the domain skew. These results suggest that FDSE succeeded in combining the advantages of two groups of existing methods, thus leading to a more generalizable representation space across different datasets.

\subsection{Generalizability to Unseen Domains}\label{sec_unseen}
We evaluate the adaptability of FDSE to unseen domains in Table \ref{table_unseen}. Each column in Table \ref{table_unseen} represents a target client that did not participate in the training process, and we evaluate the model trained on other clients on the target after adaptation. We emphasize that the model adaptation processes for all methods are label-free, as we detailed in the supplementary materials.
\textbf{- align} refers to directly evaluating the model trained by FDSE without adaptation. Our FDSE achieves comparable results to the baselines even before model adaptation. After adaptation, FDSE significantly enhances model performance for most clients, confirming FDSE's ability to generalize effectively to unseen domains.
% We verify the ability of FDSE to be adapted to unseen domains in Table \ref{table_unseen}. Each column represents the target domain client that did not join in training and we adapt the model trained on other clients to the target one for evaluation. We key that the model adaptation processes of all methods are label-free and detail them in supplementary materials.  
% % For FedAvg, we directly use the global model to make predictions. For FedBN, we first collect local statistics for 1 epoch and then evaluate the adapted model. For FDSE, we fine-tune the DSE modules and fix other parameters to minimize the consistency regularization (e.g., Sec. \ref{sec_reg}) for several epochs before evaluation.
% \textbf{- align} refers to directly using the model trained by FDSE to produce predictions without further adaption. Our FDSE can already achieve similar results with baselines before model adaption. After model adaptation, FDSE can significantly improve the model performance for most clients gainst other baselines, confirming FDSE's ability to be generalized to unseen domains.
\subsection{Impact of Hyper-parameters}
\begin{figure}
    \centering
    \includegraphics[width=\linewidth]{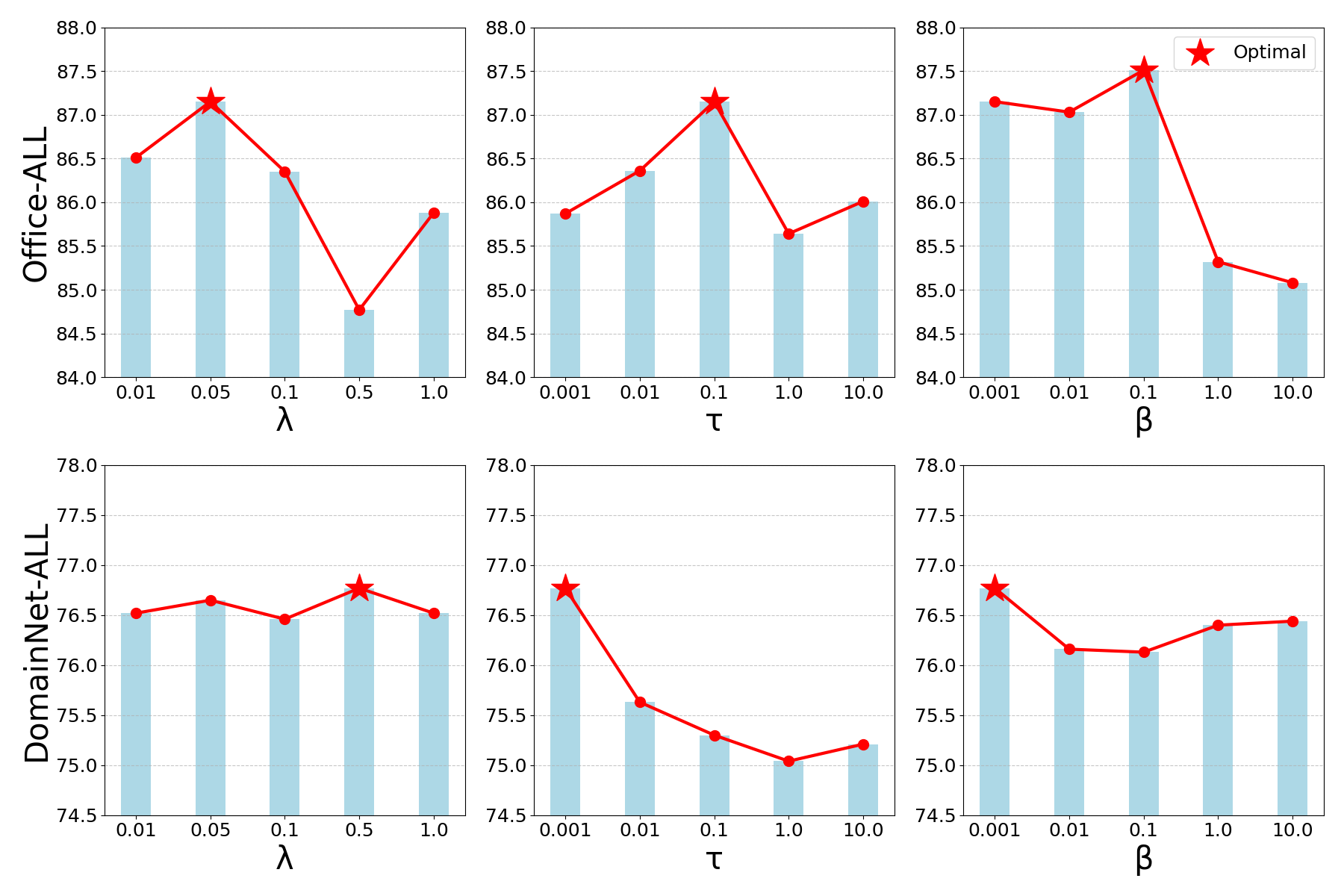}
    \caption{The impact of hyper-parameters on model performance.} \label{fig_hyper}
    \vspace{-0.2cm}
\end{figure}
\begin{table}\scriptsize
\centering
\begin{tabular}{ccc|cccccc}
\hline\hline
\multicolumn{3}{c|}{\textbf{Module}}           & \multicolumn{2}{c}{\textbf{DomainNet}} & \multicolumn{2}{c}{\textbf{Office}} & \multicolumn{2}{c}{\textbf{PACS}}  \\ 
\hline
\tiny{\textbf{A}} & \tiny{\textbf{B}} & \tiny{\textbf{C}} & \textbf{ALL} & \textbf{AVG}            & \textbf{ALL} & \textbf{AVG}                   & \textbf{ALL} & \textbf{AVG}        \\ 
\hline
 \tiny{$\times$}  &  \tiny{$\times$}   & \tiny{$\times$}  &   $74.63$    &   $ 72.19$            &  $83.94$           &    $86.13$         &   $82.46$       &  $80.72$   \\
 \tiny{$\times$}  &   \tiny{$\checkmark$}       &   \tiny{$\checkmark$}     &   $ 75.66$             &    $73.23$        &  $ 84.44$     &   $ 88.54$      &  $82.92$  &$80.97$          \\
  \tiny{$\checkmark$}  & \tiny{$\times$}    &    \tiny{$\checkmark$}   &   $76.57$           &      $ 74.32$     &   $ 85.23$           &  $ 89.25$    &  $83.40$  & $81.73$       \\
 \tiny{$\checkmark$}    &   \tiny{$\checkmark$}     &  \tiny{$\times$}  &   $76.49$    &   $74.24$         &   $86.43$           &    $ 90.38$    &  $83.40$            &  $81.78$            \\
  \hline
    \tiny{$\checkmark$}  &   \tiny{$\checkmark$}   &  \tiny{$\checkmark$}     & $ \bold{76.77}$&$\bold{74.50}$&$\bold{87.15}$&$\bold{91.58}$&$\bold{83.81}$&$\bold{82.17}$                 \\
\hline
\end{tabular}
\caption{The ablation study of the model performance of FDSE with each submodule to be respectively removed.}\label{table_ablation}
\vspace{-0.5cm}

\end{table}
\paragraph{Effect of $\lambda$.} As shown in the first column of Figure \ref{fig_hyper}, slightly increasing $\lambda$ can improves the accuracy while a too large $\lambda$ will cause performance reduction, indicating that $\lambda$ should be carefully tuned to their optimal in practice.
\vspace{-0.3cm}
\paragraph{Effect of $\beta$ and $\tau$.} 
From the last two columns of Figure \ref{fig_hyper}, the effects of both parameters exhibit a similar trend on each dataset as their values change. We attribute this similarity to the varying degrees of domain skew presented in each dataset, since small values of both $\tau$ and $\beta$ strengthen domain skew elimination. The performance is more sensitive to $\tau$ than to $\beta$, where a small $\beta$ is preferred while the optimal $\tau$ values differ, suggesting that $\tau$ should be properly tuned while $\beta$ can be fixed at a low value.

\subsection{Ablation Study}
We conduct the ablation analysis of the effectiveness of FDSE's modules in Table \ref{table_ablation}. Module A, B, and C respectively correspond to the consensus-maximization aggregation in Sec.\ref{sec_conagg}, similarity-aware aggregation in Sec.\ref{sec_simagg}, and consistency regularization in Sec.\ref{sec_reg}.  The results show that the raw architecture of FDSE has achieved comparable results with FedBN. Further, the performance will be degraded after removing each module and the optimal result appears in the full participation of each module, indicating the collaborative effectiveness of these modules.
% 0.4017857142857143 | 0.6842105263157895 | 0.8666666666666667 | 0.7241379310344828 |

% \usepackage{multirow}
% \usepackage{booktabs}

\begin{table}\small
\centering
\begin{tabular}{ccccc} 
\hline\hline
\textbf{Method} & \textbf{Num$_{\times 10^7}$} & \textbf{Comm.} & \textbf{FLOPs$_{\times 10^{10}}$} & \textbf{Time$_{train}$}  \\ 
\midrule
\textbf{FedBN}           & 1.30         & 49.52M         & 4.41           & 4.03s           \\
\textbf{Ditto}           & 1.30         & 49.52M         & 4.41           & 6.59s           \\
\textbf{FDSE}            & 0.65         & 24.87M         & 2.24           & 4.27s           \\
\bottomrule
\end{tabular}
\caption{Comparison of communication and communication costs.}\label{table_eff}
\vspace{-0.5cm}
\end{table}
\subsection{Efficiency}
 Table \ref{table_eff} compares the communication and computation costs. 
 % \textbf{Comm.} refers to the transferred model size and \textbf{FLOPs} is evaluated with batch size 50. 
 Our FDSE saves more communication efficiency per round (e.g., Comm.) and computation efficiency (e.g., FLOPs) than baseline due to its decomposition-based architecture. Besides, FDSE achieves competitive time costs of local training against FedBN, where the additional training costs of FDSE mainly come from the regularization term. 
\section{Conclusion}
In this work, we rethink the domain shift problem in FL from a hybrid view that integrates the advantages of personalization-based methods and consensus-based methods. We develop a novel framework, FDSE, to differently erase domain skew for each client while maximizing their consensus. Specifically, we efficiently formulate the model forward passing as an iterative deskewing process that extracts and then deskews features alternatively via layer decomposition. Further, we fine-grained design aggregation strategies and the regularization term to enhance the knowledge decoupling, leading to improved consistency in the representation space. We plan to extend this framework to more applications and model architectures in the future.
\section*{Acknowledgement}
The research was supported by Natural Science Foundation of China (62272403).

{
    \small
    \bibliographystyle{ieeenat_fullname}
    \bibliography{main}
}

% 修改表格编号格式
\renewcommand{\thetable}{A\arabic{table}}
% 修改图像编号格式
\renewcommand{\thefigure}{A\arabic{figure}}
\renewcommand{\thesection}{A\arabic{section}} % 修改一级章节编号
\clearpage

\setcounter{page}{1}
% \maketitlesupplementary
\begin{figure*}[!h]
    \centering
    \includegraphics[width=\linewidth]{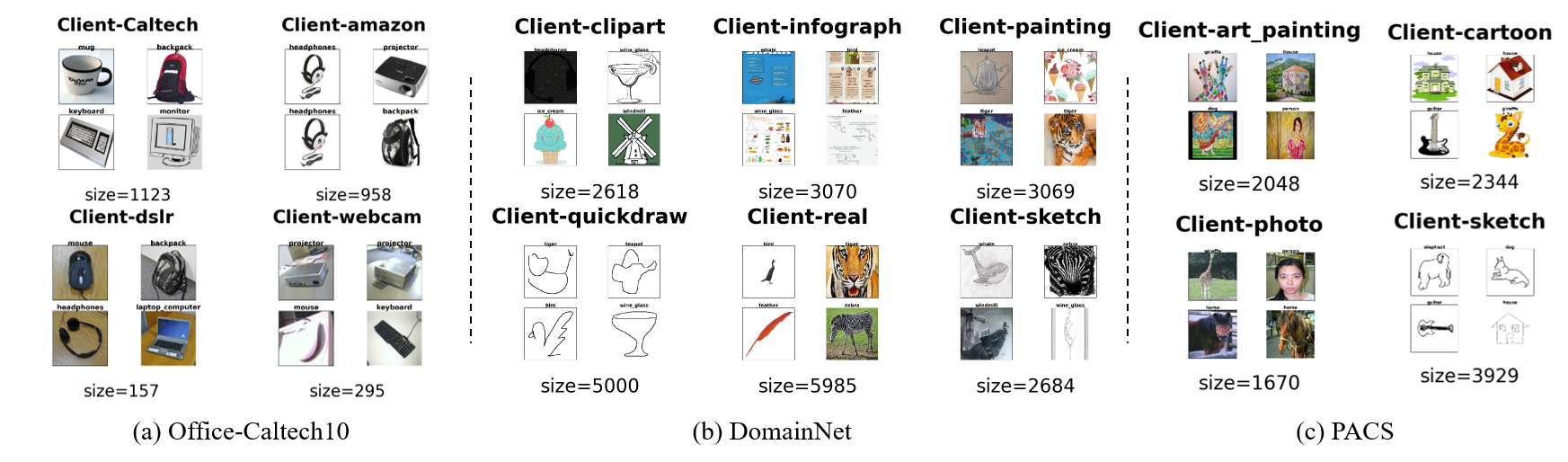}
    \caption{The visualization of each client's local data.}
    \label{fig_dataset}
\end{figure*}

\section{Experimental Details}

% \begin{figure*}[ht]
%     \centering
%     % 第一行
% \includegraphics[]{}
%     \caption{The visualization of each client's data distribution. }
%     \label{fig_dataset}
% \end{figure*}

\subsection{Datasets}
We use three popular datasets of multi-domain image classification tasks: Office-Caltech10 \cite{sun2021partialfed}, DomainNet \cite{leventidis2023domainnet}, and PACS \cite{zhou2020pacs}. The details of the three datasets are as below
\paragraph{Office-Caltech10.} Office-Caltech10 is constructed by selecting the 10 overlapping categories (e.g., backpack, bike, calculator, headphones, keyboard, laptop, monitor, mouse, mug and projector) between the Office dataset \cite{saenko2010office_raw} and Caltech256 dataset \cite{caltech256}. It contains four different domains: amazon, caltech10, dslr and webcam. These domains contain respectively 958, 1123, 295, and 157 images. 
\paragraph{DomainNet.} We follow \cite{zhou2023fedfa} to select 10 categories from the 345 categories of objects of the original dataset. The domains of this dataset include clipart, real, sketch, infograph,  painting, and quickdraw.

\paragraph{PACS.} PACS \cite{li2017pacs} consists of four domains, namely Photo (1,670 images), Art Painting (2,048 images), Cartoon (2,344 images) and Sketch (3,929 images). Each domain contains seven categories.

We follow \cite{zhou2023fedfa} to allocate each single domain's data to a client in our experiments. The visualized examples of the three datasets are respectively shown in Figure \ref{fig_dataset} (a), (b), and (c). We resize each sample into the size of $224\times224$ before feeding them into the model. We split each client's local data into training/validation/testing datasets by the ratios 0.8/0.1/0.1. The model is trained on training datasets and is selected according to its optimal performance on validation datasets. We finally report the metrics of the selected optimal model on each client's testing data.  

\begin{table}\scriptsize
\centering
\caption{Architecture of Vallina AlexNet}
\label{table_alex}
\begin{tblr}{
  cells = {c},
  vline{2} = {-}{},
  hline{1,11} = {-}{0.08em},
  hline{2} = {-}{},
}
\textbf{Layer} & \textbf{Details}                                         \\
1              & Conv2d(3, 64, 11, 4, 2), BN(64), ReLU, MaxPool2D(3,2)    \\
2              & Conv2d(64, 192, 5, 1, 2), BN(192), ReLU, MaxPool2D(3,2)  \\
3              & Conv2d(192, 384, 3, 1, 1), BN(384), ReLU                 \\
4              & Conv2d(384, 256, 3, 1, 1), BN(256), ReLU                 \\
5              & Conv2d(256, 256, 3, 1, 1), BN(256), ReLU, MaxPool2D(3,2) \\
6              & AdaptiveAvgPool2D(6, 6)                                  \\
7              & FC(9216, 1024), BN(1024), ReLU                           \\
8              & FC(1024, 1024), BN(1024), ReLU                           \\
9              & FC(1024, num\_classes)                                   
\end{tblr}
\end{table}
\begin{table}\scriptsize
\centering
\caption{Architecture of FDSE's AlexNet}
\label{table_falex}
\begin{tblr}{
  cells = {c},
  vline{2} = {-}{},
  hline{1,11} = {-}{0.08em},
  hline{2} = {-}{},
}
\textbf{Layer} & \textbf{Details}                                        \\
1              & DSEBlock(3, 64, 11, 4, 2, G=2, dw=3),~MaxPool2D(3, 2)   \\
2              & DSEBlock(64, 192, 5, 1, 2, G=2, dw=3),~MaxPool2D(3, 2)  \\
3              & DSEBlock(192, 384, 3, 1, 1, G=2, dw=3)                  \\
4              & DSEBlock(384, 256, 3, 1, 1, G=2, dw=3)                  \\
5              & DSEBlock(256, 256, 3, 1, 1, G=2, dw=3),~MaxPool2D(3, 2) \\
6              & AdaptiveAvgPool2D(6, 6)                                 \\
7              & DSEBlock(9216, 1024, 1, 1, 1, G=2, dw=1)                \\
8              & DSEBlock(1024, 1024, 1, 1, 1, G=2, dw=1)                \\
9              & FC(1024, num\_classes)                                  
\end{tblr}
\end{table}
\begin{table}\scriptsize
\centering
\caption{Architecture of DSEBlock(S,T,kernel\_size, stride, padding, G, dw)}
\label{table_dseblock}
\begin{tblr}{
  cells = {c},
  vline{2} = {-}{},
  hline{1,6} = {-}{0.08em},
  hline{2} = {-}{},
}
\textbf{Layer} & \textbf{Details}                              \\
1              & Conv2d(S, $ \lceil\text{T/G}\rceil$, kernel\_size, stride, padding),BN$_{\text{DSE}}$($ \lceil\text{T/G}\rceil$),ReLU \\
2              & Conv2d($ \lceil\text{T/G}\rceil$, T-$ \lceil\text{T/G}\rceil$, dw, 1, dw//2)  \\
3              & Concat(out$_{layer 1}$,out$_{layer 2}$)                                 \\
4              & BN$_{\text{DFE}}$(T), ReLU                                   
\end{tblr}
\end{table}
\subsection{Model Architecture}
\paragraph{Backbone.} We follow \cite{zhou2023fedfa} to use AlexNet across our experiments. The architecture of the model is as shown in Table \ref{table_alex}. The model used by FedFA has a similar architecture with Vallina AlexNet where the first five layers are respectively attached with an additional FFALayer. FDSE replaces each layer in the Vallina AlexNet with a DSEBlock as is shown in Table \ref{table_alex}, and the details of each DSEBlock are listed in Table \ref{table_dseblock}. Particularly, we follow \cite{han2020ghostnet} to preserve one identity mapping in the DSE convolution (e.g., layer 2).

\subsection{Baselines}
We consider the following baselines in this work
\begin{itemize}
    \item \textbf{Local} is a non-federated method where each client independently trains its local model;
    \item \textbf{FedAvg} \cite{mcmahan2017communication} is the classical FL method that iteratively averages the locally trained models to update the global model;
    \item \textbf{LG-FedAvg}\cite{liang2020think} is a method that jointly learns compact local representations on each device and a global model across all devices.
    \item\textbf{FedProx}\cite{li2020federated} restricts the model parameters to be close to the global ones during clients' local training to alleviate the negative impact of data heterogeneity.
    \item \textbf{Scaffold}\cite{karimireddy2020scaffold} corrects the model updating directions during model training to mitigate client drift's effects.
    \item\textbf{FedDyn}\cite{acar2021federated} maintains consistent local and global objectives during model training to avoid model overfitting on local objectives. 
    \item \textbf{MOON}\cite{li2021model} restricts the model's representation space to be close to the global ones during clients' local training.
    \item \textbf{Ditto} \cite{li2021ditto} personalizes the local model by limiting its distance to the global model for each client with a proximal term.
    \item \textbf{PartialFed}\cite{sun2021partialfed} personalizes partial model parameters to suit the global model to local distributions.
    \item \textbf{FedBN}\cite{li2021fedbn} lets BN layers be locally kept by each client without aggregation to adapt the global model to their local datasets.
    \item \textbf{FedFA}\cite{zhou2023fedfa} augments features in the intermediate layers of the model to enhance clients' consensus from the feature level.
    \item \textbf{FedHeal}\cite{chen2024fair} mitigates gradient conflicts of important model parameters to enhance clients' consensus from the model parameter level.
\end{itemize}
\begin{algorithm}[tb]
    \caption{FedBN-Adaption}
    \label{alg:2}
    \textbf{Input}:The trained model $\mathcal{M}$, the target domain's testing data $\mathcal{D}_{target}$
    \begin{algorithmic}[1]
    \FOR{ batch data $(\bold X,y) \in \mathcal{D}_{target}$}
    
    % \STATE The server checks the set of available clients $A_t$.
    % \STATE The server uses $\bold v^t$ and $\bold H$ to solve equation (16) within the maximum wall-clock time $\tau_{max}$ to obtain the sampled client set $S_t\subseteq A_t$
    \STATE the target client collects local statistics by computing $\mathcal{M}(\bold X)$
    % \STATE The server updates t he    sampling counts of clients $\bold v^{t+1}[k]\leftarrow \bold v^{t}[k] + \mathbb{I}(k\in S_t)$
    \ENDFOR
    \RETURN $\mathcal{M}$
    \end{algorithmic}
    \end{algorithm}

\begin{algorithm}[tb]
    \caption{FDSE-Adaption}
    \label{alg:3}
    \textbf{Input}:The trained model $\mathcal{M}$, the target domain's testing data $\mathcal{D}_{target}$, the number of epochs $E$, the learning rate $\eta$
    \begin{algorithmic}[1]
    \STATE  the target client freezes the gradient of trainable parameters $\theta_u$ in $\mathcal{M}$ if $\theta_u$ does not belong to any DSE modules and fixes all the statistical parameters of $\text{BN}_{\text{DFE}}$.
    \FOR {epoch $i =1,..., E $}
    \FOR{ batch data $(\bold X,y) \in \mathcal{D}_{target}$}
        
    % \STATE The server checks the set of available clients $A_t$.
    % \STATE The server uses $\bold v^t$ and $\bold H$ to solve equation (16) within the maximum wall-clock time $\tau_{max}$ to obtain the sampled client set $S_t\subseteq A_t$
    \STATE the target client computes model forward $\mathcal{M}(\bold X)$
    \STATE the target client hook DSE module's outputs $\{\bold X_{k}^{(l)}\}$
    \STATE the target client compute regularization term in Sec. 4.2
    \STATE the target client optimizes the non-frozen parameters to minimize the regularization term via gradient descent with step size $\eta$.
    % \STATE The server updates t he    sampling counts of clients $\bold v^{t+1}[k]\leftarrow \bold v^{t}[k] + \mathbb{I}(k\in S_t)$
    \ENDFOR
    \ENDFOR
    \RETURN $\mathcal{M}$
    \end{algorithmic}
    \end{algorithm}
\subsection{Hyper-parameters}
\paragraph{Common parameters.} We respectively tune the learning rate $\eta\in\{0.001, 0.01, 0.05, 0.1, 0.5\}$ by grid search for each method. We clip the gradient's norm to be no larger than 10. We run each trial for 500 communication rounds.  The batch size is fixed to $50$ and the local epochs for Domainnet, Office-Caltech10, and PACS are respectively $5$, $1$, and $5$. We decay the learning rate by the ratio $0.998$ per round. We select all the clients at each communication round like other works in cross-silo FL \cite{luo2022adapt}.
\paragraph{Algorithmic parameters.} For Ditto \cite{li2021ditto} and FedProx \cite{li2020federated}, we tune the regularization coefficient $\mu\in[0.0001, 0.001, 0.01, 0.1, 1.0]$. For MOON \cite{li2021model}, we follow its setting to set the range of the coefficient $\mu$ as $[0.1, 1.0, 5.0, 10.0]$ and fix the value of $\tau=0.5$. For FedDyn \cite{acar2021federated}, we tune the regularization coefficient $alpha\in[0.001, 0.01, 0.03, 0.1]$. For FedHeal, we tune the $\tau\in[0.1, 0.2, 0.3, 0.4, 0.5]$. For FDSE, we fix $\beta=0.001$ and only tune $\lambda\in[0.01, 0.1, 1.0], \tau\in[0.001, 0.01, 0.1, 0.5]$.
\subsection{Adapation Details}

We illustrate the details of model adaptation for each method in Sec. 5.4.  For FedAvg, we directly use the global model to make predictions on the target domain. For FedBN, we first collect local statistics for 1 epoch on the target domain's testing dataset and then evaluate the adapted model, as is shown in Algo. \ref{alg:2}. For FDSE, we fine-tune the DSE modules and fix other parameters to minimize the consistency regularization (e.g., Sec. 4.2) for several epochs before evaluation as shown in Algo. \ref{alg:3}.

% \textbf{- align} refers to directly using the model trained by FDSE to produce predictions without further adaption. Our FDSE can already achieve similar results with baselines before model adaption. After model adaptation, FDSE can significantly improve the model performance for most clients gainst other baselines, confirming FDSE's ability to be generalized to unseen domains.

\section{Additional Experiments}
\begin{figure}
\centering
    {%
        \includegraphics[width = .45\linewidth]{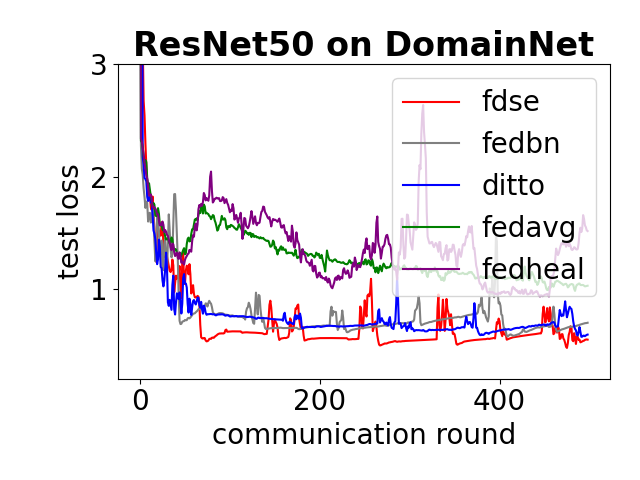}
        \includegraphics[width = .45\linewidth]{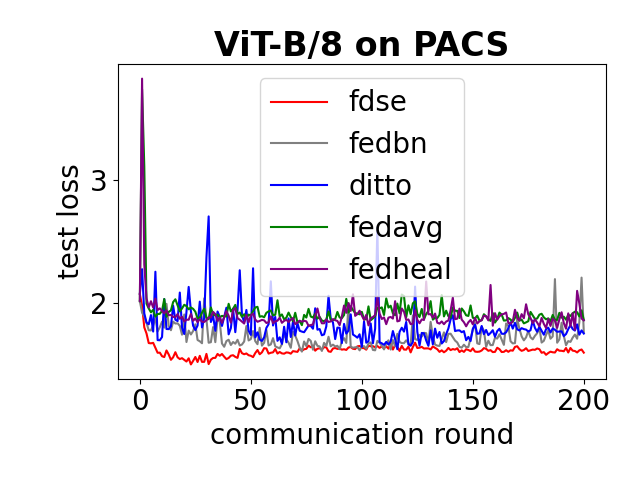}}

\caption{Testing loss curves on other model architectures.}
\label{tb_loss}
\end{figure}

\begin{table}
\scriptsize
\centering
\caption{Model performance ($\uparrow$) on other model architectures.}
\label{tb_model}
\begin{tabular}{c|cccc} 
\hline\hline
\multirow{2}{*}{\scriptsize\textbf{Method}} & \multicolumn{2}{c}{\scriptsize\textbf{DomainNet-ResNet50}} & \multicolumn{2}{c}{\scriptsize\textbf{PACS-ViT-B/8}}  \\ 
\cline{2-5}
                                 & \textbf{ALL}      & \textbf{AVG}                & \textbf{ALL}      & \textbf{AVG}           \\ 
\hline
\scriptsize\textbf{FedAvg}                  & $59.90_{\pm0.96}$ & $58.71_{\pm1.06}$           & $26.44_{\pm4.16}$ & $25.94_{\pm4.46}$      \\
\scriptsize\textbf{FedHeal}                 & $66.16_{\pm0.62}$ & $64.52_{\pm0.55}$           & $30.05_{\pm3.44}$ & $29.51_{\pm2.98}$      \\
\scriptsize\textbf{FedBN}                   & $69.36_{\pm0.29}$ & $66.99_{\pm0.53}$           & $36.67_{\pm2.01}$ & $36.22_{\pm2.86}$      \\
\scriptsize\textbf{Ditto}                   & $67.70_{\pm0.36}$ & $64.99_{\pm0.49}$           & $31.96_{\pm4.32}$ & $31.75_{\pm4.13}$      \\
\scriptsize\textbf{FDSE}                    & $\bold{72.98_{\pm0.39}}$ & $\bold{70.44_{\pm0.32}}$           & $\bold{38.24_{\pm1.69}}$     & $\bold{38.41_{\pm1.90}}$          \\
\hline
\end{tabular}
\end{table}

\begin{table}
\scriptsize
\centering
\caption{Model performance ($\uparrow$) on unseen clients. }
\label{tb_unseen2}
\begin{tabular}{c|c|ccccc} 
\hline\hline
\multicolumn{2}{c|}{\textbf{Dataset}}                               & \textbf{FedAvg} & \textbf{FedBN} & \textbf{FedDG-GA} & \textbf{FedSR} & \textbf{FDSE}  \\ 
\hline
\multirow{5}{*}{\rotatebox{90}{\textbf{Office}}} & \textbf{C}   & 51.78           & \textbf{60.71}          & 55.35             & 56.25          & 57.14          \\
                                                     & \textbf{A}   & 70.52           & 70.52          & 72.63             & \textbf{75.78}          & \textbf{75.78}          \\
                                                     & \textbf{D}   & 80.00           & 80.00          & 86.66             & 86.66          & \textbf{93.33}          \\
                                                     & \textbf{W}   & 65.61           & 55.17          & 68.96             & 69.32          & \textbf{75.86}          \\ 
\cline{2-7}
                                                     & \textbf{avg} & 66.95           & 66.60          & 70.90             & 72.00          & \textbf{75.52}          \\ 
\hline
\multirow{7}{*}{\rotatebox{90}{\textbf{DomainNet}}}        & \textbf{C}   & 62.81           & 62.56          & 62.43             & 60.75          & \textbf{65.22}          \\
                                                     & \textbf{I}   & 30.15           & 31.14          & 30.70             & 31.81          & \textbf{32.34}          \\
                                                     & \textbf{P}   & 55.53           & 57.26          & 57.04             & 56.18          & \textbf{59.32}          \\
                                                     & \textbf{O}   & 48.86           & 53.06          & 48.26             & 52.13          & \textbf{55.00}          \\
                                                     & \textbf{R}   & 59.74           & 63.17          & 59.85             & 64.15          & \textbf{64.28}          \\
                                                     & \textbf{S}   & 58.92           & 62.46          & 58.92             & 58.55          & \textbf{65.28}          \\ 
\cline{2-7}
                                                     & \textbf{avg} & 52.66           & 54.94          & 52.87             & 53.92          & \textbf{56.91}          \\
\hline
\end{tabular}
\end{table}
\subsection{Other Model Architecture}

 We have studied the effectiveness on relatively large models in Table \ref{tb_model}. We replace the last operator of each layer (i.e., ResNet50's block and Vit-B/8's feedforward layer) with DSE module. FDSE consistently outperforms baselines (e.g., Table \ref{tb_model}) and exhibits faster convergence speed (e.g., Figure \ref{tb_loss}).

\subsection{Additional Baselines of generalizabilityy}
We compare FDSE with the additional baselines \cite{fedsr, feddgga} for unseen clients in Table \ref{tb_unseen2}. FDSE outperforms all baselines, which we attribute to the additional adaptation steps.

% \label{sec:rationale}
% % 
% Having the supplementary compiled together with the main paper means that:
% % 
% \begin{itemize}
% \item The supplementary can back-reference sections of the main paper, for example, we can refer to \cref{sec:intro};
% \item The main paper can forward reference sub-sections within the supplementary explicitly (e.g. referring to a particular experiment); 
% \item When submitted to arXiv, the supplementary will already included at the end of the paper.
% \end{itemize}
% % 
% To split the supplementary pages from the main paper, you can use \href{https://support.apple.com/en-ca/guide/preview/prvw11793/mac#:~:text=Delete%20a%20page%20from%20a,or%20choose%20Edit%20%3E%20Delete).}{Preview (on macOS)}, \href{https://www.adobe.com/acrobat/how-to/delete-pages-from-pdf.html#:~:text=Choose%20%E2%80%9CTools%E2%80%9D%20%3E%20%E2%80%9COrganize,or%20pages%20from%20the%20file.}{Adobe Acrobat} (on all OSs), as well as \href{https://superuser.com/questions/517986/is-it-possible-to-delete-some-pages-of-a-pdf-document}{command line tools}.
% {
%     \small
%     \bibliographystyle{ieeenat_fullname}
%     \bibliography{main}
% }

\end{document}